# Adpositional Supersenses for Mandarin Chinese

Yilun Zhu
*Georgetown University*, yz565@georgetown.edu

Yang Liu
*Georgetown University*, yl879@georgetown.edu

Siyao Peng
*Georgetown University*, sp1184@georgetown.edu

Austin Blodgett
*Georgetown University*, ajb341@georgetown.edu

Yushi Zhao
*Georgetown Unviersity*, yz521@georgetown.edu

*See next page for additional authors*

# Adpositional Supersenses for Mandarin Chinese

**Authors**
Yilun Zhu, Yang Liu, Siyao Peng, Austin Blodgett, Yushi Zhao, and Nathan Schneider

# Adpositional Supersenses for Mandarin Chinese

Yilun Zhu  Yang Liu  Siyao Peng  Austin Blodgett  Yushi Zhao  Nathan Schneider

Georgetown University

{yz565, yl879, sp1184, ajb341, yz521, nathan.schneider}@georgetown.edu

## 1 Introduction

Adpositions, though belonging to a closed functional category, can contribute significantly to meaning. Schneider et al. (2018b) proposed an annotation scheme called Semantic Network of Adposition and Case Supersenses (SNACS), which includes 50 supersense labels (LOCUS, TOPIC, etc.). Unlike other approaches to semantic tagging/role labeling, SNACS incorporates the *construal analysis* (Hwang et al., 2017) wherein the lexical semantic contribution of an adposition token (its **function**) is distinguished and may diverge from the underlying relation in the surrounding context or **scene**. For instance, (1) blends the domains of emotion (principally reflected in *care*, which licenses a STIMULUS), and cognition (principally reflected in *about*, which often marks non-emotional TOPICs). The token is therefore annotated with both supersenses; we use the notation SCENEROLE↝FUNCTION:

(1)    I care **about:STIMULUS↝TOPIC** you.[1]

The SNACS scheme was developed for English and tested for annotation and automatic disambiguation on English corpora. Though other languages were taken into consideration in designing SNACS, no serious annotation effort has been undertaken to confirm empirically that it generalizes to other languages. Here, we adapt SNACS annotation to Mandarin Chinese and demonstrate that the same supersense categories are appropriate for Chinese adposition semantics. We annotate 20 chapters of *The Little Prince* in Chinese, giving an English-Chinese parallel corpus to examine similarities and differences in prepositional construal between the two languages.

## 2 Adposition Criteria

The first challenge is in determining which words (and multiword expressions) qualify as meriting SNACS supersenses. For example, *coverbs* and *localizers* are categories in Chinese grammar that bear some relationship to adpositions, though their precise classification is controversial. In (2), *xuéshù* (i.e. 'academia') is surrounded by a coverb *zài* and a localizer *shàng*.

(2)   tā    **zài:LOCUS**   xuéshù
      3SG   P:at           academia
      **shàng:TOPIC↝LOCUS**   yǒusuǒzuòwéi.
      LC:on-top-of              successful
      'He succeeded in academia.'

### 2.1 Coverbs

Coverbs usually precede the main predicate of the clause and introduce an NP argument to it (Li and Thompson, 1974). In (2), the noun surrounded by the coverb *zài* (functioning as a preposition) and the localizer *shàng* precedes the predicate *yǒusuǒzuòwéi*. In some cases, coverbs can also occur as predicates. For example, the coverb *zài* heads the predicate phrase in (3), different from those occurring in a modifier position. In this project, we annotate all coverbs only when they occur pre-verbally, echoing the view that coverbs modify events introduced by the predicates, rather than establishing multiple events in a clause (Hui, 2012). Therefore, lexical items such as *zài* in (3) are not annotated.

(3)   nǐ   yào  de   yáng  jiù   **zài** lǐmiàn.
      2SG  want DE   sheep RES   at    inside
      'The sheep you want is in the box.'
      (zho_lpp_1943.92)

### 2.2 Localizers

Localizers are words that follow a noun phrase to refine its semantic relation. E.g., *shàng* in (2) denotes a contextual meaning, 'in a particular topic', whereas the co-occurring coverb *zài* only conveys a generic location. It is unclear whether localizers are syntactically postpositions, but we annotate all localizers because of their semantic significance.

Though coverbs frequently cooccur with localizers, the combinations are somewhat productive, so we treat them as separate targets for SNACS annotation. Thus, *zài* and *shàng* receive LOCUS and TOPIC↝LOCUS respectively in (2).

## 3 Supersense Applicability

For the most part, we found that the SNACS supersenses developed for English could be applied to Chinese. However, we identified a number of dif-ferent construals that were frequent in Chinese but rare or unattested in English. A couple of examples are noted below.

---

[1] Throughout this abstract, we only bold and label supersense for adpositions that are relevant to the discussion.

### 3.1 EXPERIENCER as Function

In English, some supersenses, such as EXPERIENCER, do not seem to have any prototypical adpositions (Schneider et al., 2018a). In (4), the scene role EXPERIENCER is expressed through the preposition *to* and construed as GOAL, which highlights the abstract destination of the *air of truth*. This reflects the basic meaning of *to*, which denotes a path towards a goal (Bowerman and Choi, 2001). In contrast, the lexicalized combination of the preposition *duì* and the localizer *láishuō* in (5) jointly establish a functionality to introduce the mental state of the experiencer, denoting the meaning 'to someone's regard'. The high frequency of such combination in the annotated corpus (11 occurrences) indicates that EXPERIENCER does have a prototypical adposition in Chinese.

(4) **To:EXPERIENCER⤳GOAL** those who understand life, that would have given a much greater air of truth to my story. (en_lpp_1943.185)

(5) [**duì:EXPERIENCER** [dǒngdé shēnghuó
    P:to           know-about life
    de rén]    **láishuō:EXPERIENCER**],
    DE people  LC:one's-regard
    zhèyàng shuō jiù  xiǎndé zhēnshí
    this-way tell RES seems  real
    'It looks real to those who know about life.' (zh_lpp_1943.185)

### 3.2 Same Scene Role, Different Function

Both English *to* and Chinese *duì* have RECIPIENT as the scene role. In (6), GOAL is labelled as the function of *to* because it indicates the completion of the "saying" event.[2] In Chinese, *duì* has the function label DIRECTION provided that *duì* highlights the orientation of the message uttered by the speaker as in (7). Even though they express the same scene role in the parallel corpus, their lexical semantics still requires them to have different functions.

(6) You would have to say **to:RECIPIENT⤳GOAL** them: "I saw a house that costs $20,000." (en_lpp_1943.172).

---

[2]The prototypical function of *to* indicates telic motion events. Telicity, however, is not required to DIRECTION.

(7) (nǐ) bìxū [**duì:RECIPIENT⤳DIRECTION**
    2SG must P:to
    tāmen] shuō: "wǒ kànjiàn le  yí dòng
    3PL    say   1SG see   ASP one CL
    shíwàn fǎláng de fángzi."
    10,000 franc   DE house
    'You must tell them: "I see a house that costs 10,000 francs." ' (zh_lpp_1943.172).

### 3.3 Unproductivity of Function

Throughout the annotated data, Chinese adpositions have relatively limited functions compared to English. For example, in English, the functions of *in* include LOCUS, TIME, MANNER, as well as TOPIC as in (8) and (9). In Chinese, however, LOCUS is the only function label for the paralleled localizer *shàng*, and the scene role is expressed through the construal TOPIC⤳LOCUS as in (10).

(8) **In:TOPIC** certain more important details I shall make mistakes. (en_lpp_1943.201)

(9) I should have liked to begin this story **in:MANNER** the fashion of the fairy-tales. (en_lpp_1943.183)

(10) wǒ hěn kěnéng [**zài:LOCUS** [mǒuxiē
     1SG very probably P:at        some
     zhòngyào de xìjié]
     important DE detail
     **shàng:TOPIC⤳LOCUS**] huà  cuò
     LC:on-top-of          draw be-mistaken
     le.
     ASP
     'I probably made mistakes on some important details.' (zh_lpp_1943.201)

## 4 Corpus Annotation and Evaluation

20 chapters of *The Little Prince* have been preprocessed by Stanford Word Segmenter (Chang et al., 2008)[3] using 'ptb'-mode with subsequent manual corrections following the Penn Chinese Treebank guidelines (Xia, 2000).

**Corpus annotation.** With our Chinese-specific guidelines, we annotated 20 chapters of *The Little Prince* in Chinese consisting of 13,000+ tokens. 14 chapters were annotated jointly by three native Chinese speakers, all of whom had received advanced training in theoretical and computational linguistics. Among the 602 adpositions we annotated, 40 types of construals were identified, with 24 of the 50 supersenses appearing as scene roles and 23 as functions.

**Inter-annotator agreement.** As a preliminary evaluation of the reliability of the adapted scheme for Chinese, we conducted an agreement study on six chapters (Ch. 15–20), including 111 adpositions. Raw agreement was .92 on scenes, .95 on functions, and .90 on role+function combinations. Average pairwise Cohen's kappa was .90 on scene roles, .93 on functions and .88 on role+function combinations, indicating strong agreement.

---

[3]https://nlp.stanford.edu/software/segmenter.html

## 5 Cross-lingual Correspondence

We compared chapters 1, 4, and 5 of *The Little Prince* that are annotated both in English[4] and Chinese,[5] and found that the inventory of supersenses captures the cross-linguistic similarities of adpositional semantics between the two languages.

### 5.1 Alignment

Among the 256 English and 141 Chinese adpositions in the three chapters of *The Little Prince*, 71 are manually aligned based on two criteria: (i) the matching mentions in English and Chinese both appear as NP constituents; and (ii) both NP constituents are governed by adpositions. For instance, the coverb and the localizer together in (11) match the adposition in (12) because the constituents governed by the adpositions in both English and Chinese are semantically equivalent to each other.

(11) ... **zài:**LOCUS⤳LOCUS yì běn miáoxiě
... P:at one CL describe
yuánshǐ sēnlín de míngjiào "zhēnshí de
primeval forest DE call true DE
gùshì" de shū **zhōng:**LOCUS⤳LOCUS...
story DE book LC:in
'In a book about primeval forest called True Stories.' (zh_lpp_1943.2)

(12) ... **in:**LOCUS⤳LOCUS a book, called True Stories from Nature, about the primeval forest. (en_lpp_1943.2)

73% of the aligned adpositions share the same scene role and 51% share the same function. This result matches the principles of the construal analysis in which scene roles capture the contextual usage (Hwang et al., 2017) and thus are more frequently matched in a bilingual parallel corpus.

### 5.2 Applicability of Contrual Analysis in Chinese

We compared scene role and function annotations in three chapters in English and Chinese and found that most adpositions have identical scene and function. As shown in Table 1, for 70% of annotated English tokens, both scene role and function receive the same supersense, and the same holds for 86% of Chinese tokens.[6]

| *Role vs. Function:* | **Same** | **Total** |
|---|---|---|
| English | 178 (70%) | 256 |
| Chinese | 121 (86%) | 141 |

Table 1: Comparing the scene role and function annotations for adpositions in Ch. 1, 4, and 5.

Though in both languages the lexical semantics of adpositions (i.e., function supersense) and their contextual usage (i.e., scene role supersense) usually have the same type, 14% and 30% of adposition tokens have different usages in context in Chinese and English respectively. In the three chapters of the Chinese translation, 10 distinct adpositions are represented in the 14% of tokens whose scene role does not agree with the function, versus 34 distinct adpositions in the remaining tokens (86%). Though scene matches function in many Chinese adpositions, some particular adpositions (e.g. *zhōng*) vary their lexical semantics to different usages (e.g. CIRCUMSTANCE and MANNER). A non-construal analysis would not capture the richness of contextual usage in some Chinese adpositions. These observations indicate that the construal analysis should be applied not only in English but also in Chinese.

### 5.3 Analysis by Subhierarchy

As in Figure 1, all 50 supersenses in SNACS are categorized into three non-overlapping subhierarchies: Circumstance (CIRC), Participant (PART), and Configuration (CONF) (Schneider et al., 2018b). Circumstance usually provides non-core information of an event, Participant involves arguments of an event, and Configuration builds up relations between two entities. The data between English and Chinese reveals the similarities that the scene role is more likely to fall into Participant if it does not match the function in the same subhierarchy.

We turn now to tokens where the scene role and function diverge significantly, i.e., are categorized in different subhierarchies. There are 38 such "cross-hierarchy" tokens in English and 14 in Chinese—the breakdown by subhierarchy appears in Table 2. Of these, overwhelmingly the scene role comes from the PART subhierarchy, such as TOPIC in (10). The distribution reveals the tendency that if adpositions are used across subhierarchies, the context favors them to be the arguments of an event in both languages. This observation represents similarities between the two languages at the subhierarchy level and seemingly restricts scenes to agree with functions within the subhierarchy. This commonality between English and Chinese demonstrates the practicality of extending SNACS to Chinese.

---

[4] English annotations are provided by Schneider et al. (2018b). The source corpus was already sentence-aligned (Abstract Meaning Representation: *The Little Prince* corpus, version 1.6, https://amr.isi.edu/download.html).

[5] 256 adposition tokens are annotated in English using 64 types of construals, versus 141 tokens and 30 types of construals in Chinese.

[6] In both languages, special labels such as DISCOURSE are included in the total number of adpositions.

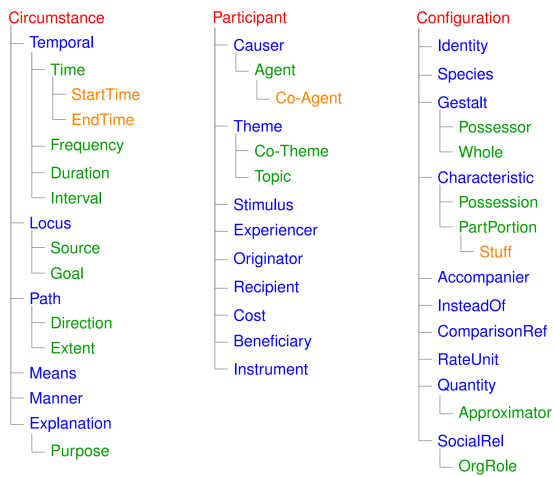

Figure 1: SNACS hierarchy of 50 supersenses.

| Func \ Scene | CIRC | PART | CONF | Total |
|---|---|---|---|---|
| CIRC | 88 ǀ 68 | 22 ǀ 12 | 4 ǀ 2 | 114 ǀ 82 |
| PART | 0 ǀ 0 | 50 ǀ 48 | 0 ǀ 0 | 50 ǀ 48 |
| CONF | 3 ǀ 0 | 9 ǀ 0 | 71 ǀ 10 | 83 ǀ 10 |
| Total | 91 ǀ 68 | 81 ǀ 60 | 75 ǀ 12 | 247 ǀ 140 |
| | | | Diagonal | 209 ǀ 126 |
| | | | Off-diagonal | 38 ǀ 14 |

Table 2: Distribution of cross-subhierarchy construals. Counts are notated as English ǀ Chinese.

Furthermore, adpositions with CONF supersense are more frequent in English than in Chinese. On the diagonal of Table 2, only 7% (10 out of 140) of adpositions in Chinese involve the CONF subhierarchy as both role and function, versus 28% (71 out of 247) in English, indicating that relationships between entities are not usually expressed by adpositions in Chinese. Though the lexical semantics of Chinese adpositions has lower tendency to fall into Configuration, this subhierarchy is still needed to describe some of the Chinese adpositions.

## 6 Conclusion

We have adapted SNACS to Mandarin Chinese, having developed new guidelines for phenomena not present in English and annotated 20 chapters of *The Little Prince*, with high interannotator agreement. The parallel corpus substantiates the applicability of construal analysis in Chinese and gives insight into the differences in construals between adpositions in two languages. The corpus can further support automatic disambiguation of adpositions in Chinese, and the common inventory of supersenses between the two languages can potentially serve cross-linguistic tasks such as machine translation.

## Acknowledgments

We would like to thank Nathan's Excellent Research Team (NERT), Case and Adposition Representation for Multi-Lingual Semantics (CARMLS) group, and two anonymous reviewers for valuable comments on previous versions of this work.